\documentclass{article}

\usepackage[preprint]{neurips_2026}

\usepackage[utf8]{inputenc} 
\usepackage[T1]{fontenc}    
\usepackage{hyperref}       
\usepackage{url}            
\usepackage{booktabs}       
\usepackage{nicefrac}       
\usepackage{microtype}      
\usepackage{xcolor}         
\usepackage{enumitem}       
\usepackage{amsmath}
\usepackage{cleveref}
\usepackage{multirow}
\usepackage{pifont}
\usepackage{graphicx}
\usepackage{amssymb}   
\usepackage{algorithm}
\usepackage{algorithmic}
\usepackage{tikz}
\usepackage{wrapfig}
\usepackage{amsthm}
\newtheorem{proposition}{Proposition}
\usetikzlibrary{arrows.meta}

\title{Interpretability Is Not Model-Specific}

\author{
  Marcus Armstrong \And Navid Ayoobi \And Pradham Mummaleti \And Alexander Chulzhanov \And Arjun Mukherjee \\
  Department of Computer Science \\
  University of Houston \\
  Houston, TX 77204 \\
  \texttt{\{miarmstr, nyoobi, pmummale, achulzha\}@cougarnet.uh.edu, amukher6@central.uh.edu}
}

\begin{document}

\maketitle

\begin{abstract}
Interpretability tools are usually built for one model at a time: a refusal direction, a truthfulness probe, or a steering vector is found with labeled data inside the model it explains. We ask whether such tools can be carried to other models through a map between residual streams fit only on unlabeled prompts. We transfer a refusal direction, tested causally by addition and ablation, and two read-out tools, a correctness direction and a sentiment probe, across 110 pairs of models from five families and 1B to 72B parameters and between base and instruction-tuned models, with controls for specificity and five seeds. How much transfers depends on what the map was fit on. A map fit on ordinary questions recovers 54\% of the target's own refusal steering and 16\% to 77\% of its correctness gain; adding unlabeled prompts from the region where the tool operates raises these to 102\% and 88\% to 98\%, while control directions recover at most 1\%. We prove that a linear map behaves this way whenever two models correspond linearly on the data it covers, and a label-free coverage score derived from this result tracks recovery across more than 16,000 runs and explains earlier reports that maps between models fail across domains. Carried read-out tools outperform anything the target's unlabeled representations yield. Interpretability results, once found, can be reused across models at the cost of a few unlabeled prompts.
\end{abstract}

\section{Introduction}
\label{sec:intro}

Interpretability research increasingly produces tools that act on a model's internal activations. A single direction in the residual stream can mediate refusal \citep{arditi2024refusal}, separate true from false statements \citep{burns2023discovering, marks2024geometry}, or score candidate answers by correctness \citep{armstrong2026correctness}, and adding or removing such directions steers behavior \citep{turner2023activation, zou2023representation, rimsky2024steering}. Each tool is built for one model, usually from labeled examples, and rebuilt from scratch for the next, so a finding about one model's internals provides no evidence about any other.

Several lines of work suggest that rebuilding may be unnecessary. Independently trained networks can be joined by simple learned maps \citep{lenc2015understanding, bansal2021revisiting}, the representations of different models have been argued to converge \citep{huh2024platonic}, and steering interventions have been carried between small models through maps trained on task data \citep{oozeer2025activation}. The evidence is mixed, however. Directions mapped across model families often change behavior no more than random directions do \citep{syed2026actionable}, and maps fit on one domain fail on another with negative $R^2$ \citep{armstrong2026thinking}. It remains unclear when a tool carried across models still works, and why it sometimes fails.

\begin{figure}[t]
  \centering
  \begin{tikzpicture}[
      font=\small,
      model/.style={draw, rounded corners=3pt, minimum width=2.6cm, minimum height=1.25cm, align=center, fill=gray!8},
      mapbox/.style={draw, rounded corners=3pt, minimum width=3.0cm, minimum height=1.25cm, align=center, fill=blue!6},
      data/.style={draw, dashed, rounded corners=2pt, align=center, inner sep=3pt, font=\footnotesize, minimum width=2.4cm},
      arr/.style={-{Stealth[length=2.2mm]}, thick},
      lab/.style={font=\footnotesize, align=center}
    ]
    \node[model] (src) at (0,0) {Source model\\[1pt]{\footnotesize tool $r_S$ built}\\[-1pt]{\footnotesize with source labels}};
    \node[mapbox] (map) at (4.5,0) {Map $M$\\[1pt]{\footnotesize ridge regression}\\[-1pt]{\footnotesize between residual streams}};
    \node[model] (tgt) at (9.0,0) {Target model\\[1pt]{\footnotesize applies $M(r_S)$}\\[-1pt]{\footnotesize no target labels}};
    \draw[arr] (src) -- node[lab, above] {$r_S$} (map);
    \draw[arr] (map) -- node[lab, above] {$M(r_S)$} (tgt);
    \node[data] (gen) at (2.75,1.85) {General text\\{\scriptsize nine question domains}};
    \node[data] (task) at (6.3,1.85) {Unlabeled task prompts\\{\scriptsize 16 to 256}};
    \draw[arr, gray] (gen.south) -- ([xshift=-7mm]map.north);
    \draw[arr, gray] (task.south) -- ([xshift=7mm]map.north);
    \node[lab, gray!70!black] at (4.5,2.65) {paired activations used to fit $M$};
    \node[lab, anchor=north] at (0,-0.8) {Tools: refusal steering,\\correctness scoring,\\sentiment classification};
    \node[lab, anchor=north] at (9.0,-0.8) {Compared with the target's\\own tool $r_T$ and with\\three control directions};
  \end{tikzpicture}
  \caption{Overview. A tool built on a source model is carried into a target model through a linear map fit on paired activations, and applied to the target without any of the target's labels. Transfer is measured as the share of the target's own effect that the carried tool reproduces. The map's training data decides how much transfers: fit on general text alone, the carried refusal direction recovers 54\% of the native effect, and adding 16 unlabeled harmful and 16 unlabeled harmless prompts raises recovery to 102\%. The carried tool $\hat r_T$ is rebuilt from the mapped representations of the prompts that define $r_S$; for a difference of means this equals the map's linear part applied to $r_S$, up to renormalization.}
  \label{fig:overview}
\end{figure}
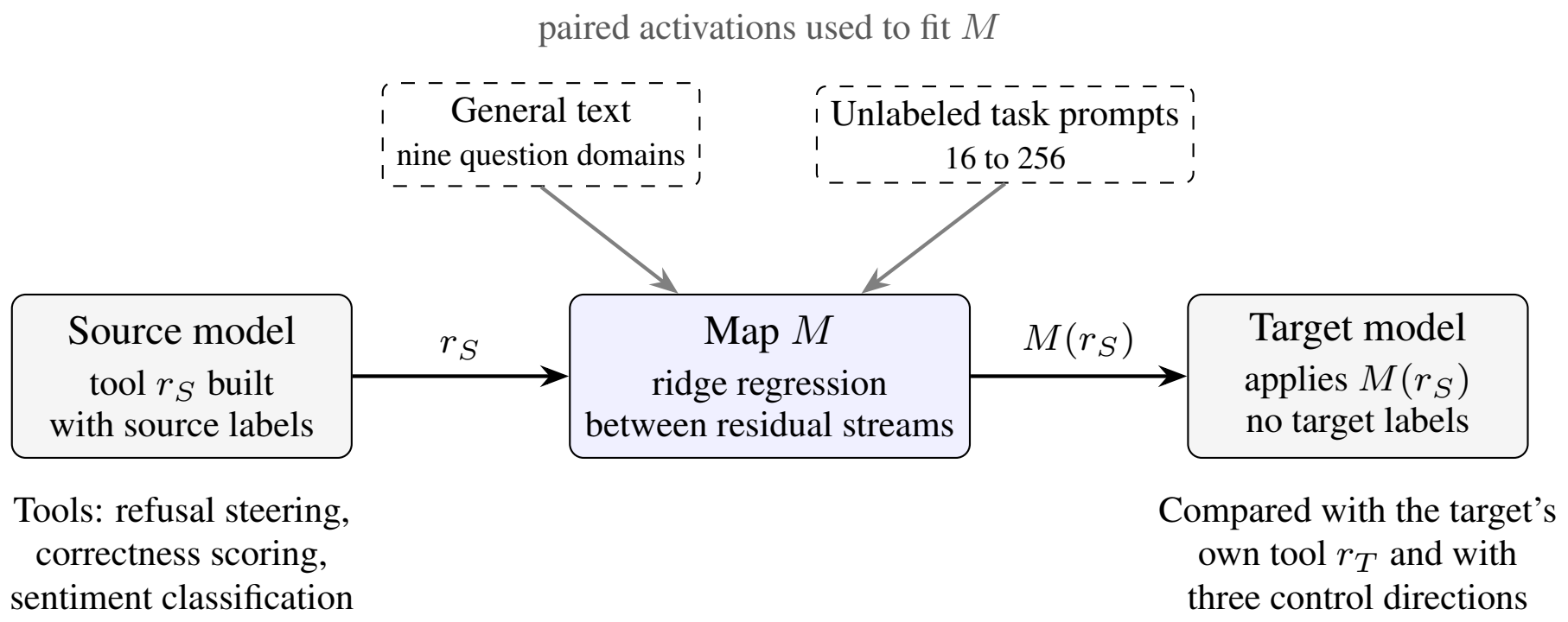

We study this across 11 instruction-tuned models from five families, from 1B to 72B parameters, and all 110 ordered pairs among them. For each pair we fit a ridge regression map from the source's residual stream to the target's on unlabeled prompts, carry a tool built on the source through it, and apply the result to the target without any target labels (Figure~\ref{fig:overview}). We transfer a refusal direction that steers generation, a correctness direction that scores answer candidates, and a sentiment probe, and compare each with the target's own tool and with random, orthogonal, and mapped-random controls.

Transfer is specific: controls recover at most 1\% of the native refusal effect and leave correctness and sentiment at or below chance. How much transfers depends on the map's training data. A map fit on ordinary questions recovers 54\% of the target's own refusal steering and 16\% to 77\% of its correctness gain. Adding unlabeled prompts from the region where the tool operates raises recovery to 102\% and 88\% to 98\%, brings the sentiment probe to the accuracy of the target's own, and carries over between base and instruction-tuned models; for refusal, 16 prompts per class suffice. We show that this is what a linear map must do: if two models correspond linearly where the map has data, the carried tool departs from the native one only through the part of the tool the data leave uncovered, and a label-free coverage score built from this result tracks recovery across more than 16,000 runs. The value of transfer depends on the tool. A refusal direction can be rebuilt in the target from the same prompts, so its transfer is evidence that the mechanism corresponds, whereas carried correctness and sentiment tools outperform any tool the target's unlabeled representations yield.

Our contributions are:
\begin{itemize}
  \item A test of cross-model transfer for three interpretability tools over 110 model pairs from 1B to 72B parameters and between base and instruction-tuned models, with specificity controls and five seeds (Sections~\ref{sec:setup} and~\ref{sec:specific}).
  \item A coverage account of transfer: a proposition bounding the error of a carried tool by the share of it the map's data leave uncovered, a label-free coverage score that tracks recovery, and an explanation of earlier cross-domain mapping failures (Section~\ref{sec:coverage}).
  \item A comparison with building tools in the target, showing where carrying replaces target labels, and a label-free measure that ranks sources (Section~\ref{sec:predict}).
\end{itemize}

\section{Related work}
\label{sec:related}

\paragraph{Representation similarity across models.}
Neural representations have been compared with similarity indices such as CKA \citep{kornblith2019similarity} and with learned maps that stitch one network into another \citep{lenc2015understanding, bansal2021revisiting}. Relative representations compare latent spaces of different models without a learned map \citep{moschella2023relative}, and the Platonic Representation Hypothesis argues that models trained on different data and objectives converge toward a shared representation \citep{huh2024platonic}. For language models, an affine map between final hidden states preserves much of their downstream behavior \citep{gorbett2026ventriloquist}, sparse autoencoder features recur across models \citep{lan2024sparse}, and information shared across models can be decoded from each while exerting little causal influence on its predictions \citep{usama2026convergence}. Maps between models have also been reported to fail across reasoning domains \citep{armstrong2026thinking}. We measure similarity through the function of transferred tools, and we trace this cross-domain failure to the data used to fit the map.

\paragraph{Transferring interventions between models.}
\citet{oozeer2025activation} transfer steering vectors for backdoor removal and refusal through autoencoder and affine maps trained on each task's own data, in models of up to 3B parameters. \citet{syed2026actionable} map a misalignment direction across four model families with ridge regression and find that the mapped direction changes behavior, while random and orthogonal control directions change it about as much. \citet{agarwal2026cross} report that cross-architecture steering succeeds for most concepts in models above roughly 1.7B parameters and degrades below that size. These studies leave open what makes transfer succeed. We fit ridge regression maps, vary their training data, never use target labels, and test up to 110 pairs from 1B to 72B parameters against random, orthogonal, and mapped-random controls with five seeds.

\paragraph{Directions as interpretability tools.}
The linear representation hypothesis holds that many concepts correspond to directions in activation space \citep{park2024linear}. Linear probes read such directions out \citep{alain2017understanding, belinkov2022probing}, and steering methods add or remove them to change behavior \citep{turner2023activation, li2023inference, zou2023representation, rimsky2024steering}. Refusal is mediated by a single direction in many chat models \citep{arditi2024refusal}, although refusal in broader contexts can require several \citep{wang2026steering}. Correctness and truthfulness are also linearly recoverable from hidden states \citep{burns2023discovering, marks2024geometry, orgad2025llms, armstrong2026correctness}, and how reliably a steering vector works depends on how coherently its training data represents the behavior \citep{braun2025understanding}. This work builds and evaluates each direction within a single model. We ask when such a direction remains valid in another.

\section{Setup}
\label{sec:setup}

\begin{table}[t]
  \centering
  \caption{Models and the native effect of each tool. All rates and accuracies are percentages. Refusal: baseline refusal on harmful prompts, and refusal on harmless prompts when the target's own direction is added at depth $d$ and working strength $\alpha^*$. Correctness: zero-shot and native-direction accuracy. Sentiment: native-direction accuracy (chance is 50\%). Native values are means over five seeds; Mistral-7B's native refusal uses one. The two largest models run in 4-bit precision. Entries marked $\dagger$ fail the inclusion criterion for that tool (Section~\ref{sec:setup}).}
  \label{tab:models}
  \small
  \setlength{\tabcolsep}{4.5pt}
  \begin{tabular}{@{}lrrccrrccr@{}}
    \toprule
    & & & \multicolumn{4}{c}{Refusal} & \multicolumn{2}{c}{Correctness} & Sentiment \\
    \cmidrule(lr){4-7} \cmidrule(lr){8-9} \cmidrule(l){10-10}
    Model & Params & Layers & $d$ & $\alpha^*$ & Base & Native & ARC & TruthfulQA & Native \\
    \midrule
    Llama-3.2-1B  & 1.0B  & 16 & 0.50 & 0.4 & 73 & 53 & $37 \to 34^{\dagger}$ & $30 \to 73$ & 92 \\
    Qwen2.5-1.5B  & 1.5B  & 28 & 0.50 & 0.2 & 91 & 77 & $45 \to 49^{\dagger}$ & $29 \to 66$ & 67 \\
    Gemma-2-2B    & 2.6B  & 26 & 0.50 & 0.4 & 94 & 100 & $52 \to 65$ & $39 \to 72$ & 94 \\
    Phi-3-mini    & 3.8B  & 32 & 0.50 & 0.4 & 94 & 92 & $52 \to 77$ & $36 \to 73$ & 80 \\
    Qwen2.5-7B    & 7.0B  & 28 & 0.50 & 0.4 & 89 & 57 & $52 \to 78$ & $40 \to 71$ & 94 \\
    Mistral-7B    & 7.0B  & 32 & 0.50 & 0.8 & $9^{\dagger}$ & 20 & $57 \to 69$ & $39 \to 74$ & 91 \\
    Llama-3.1-8B  & 8.0B  & 32 & 0.50 & 0.8 & 86 & 99 & $53 \to 69$ & $34 \to 78$ & 90 \\
    Gemma-2-9B    & 9.0B  & 42 & 0.50 & 0.4 & 95 & 94 & $64 \to 85$ & $43 \to 82$ & 95 \\
    Gemma-2-27B   & 27B   & 46 & 0.50 & 0.1 & 89 & 50 & $64 \to 89$ & $47 \to 85$ & 96 \\
    Llama-3.1-70B & 70B   & 80 & 0.50 & 0.8 & 89 & 71 & $58 \to 82$ & $41 \to 81$ & 92 \\
    Qwen2.5-72B   & 72B   & 80 & 0.75 & 0.8 & 64 & 100 & $58 \to 93$ & $45 \to 84$ & 93 \\
    \bottomrule
  \end{tabular}
\end{table}

\paragraph{Models and representations.}
We use 11 instruction-tuned models from five families, from 1B to 72B parameters (Table~\ref{tab:models}); the two largest run in 4-bit precision. A prompt's representation is the residual stream at its last token after block $\lfloor dL \rceil$ of a model with $L$ blocks, normalized to unit length. Refusal uses chat-formatted prompts at $d = 0.5$, except for Qwen2.5-72B, whose own direction works only at $d = 0.75$ (Appendix~\ref{app:exclusions}). Correctness and sentiment use $d = 0.75$ with the templates \texttt{Question: \{q\} Answer: \{a\}} and \texttt{Review: \{t\} Sentiment:}.

\paragraph{Maps.}
A map is a ridge regression with intercept from source to target representations of the same prompts, with its penalty chosen by exact leave-one-out error (Appendix~\ref{app:maps}). We vary its training data. \emph{General} uses 5,400 ordinary questions, 600 from each of nine domains. \emph{General + task} adds unlabeled prompts from the region where the tool operates: 256 harmful and 256 harmless instructions, the answer candidates of 200 questions per dataset, or 256 reviews. \emph{Task only} uses those prompts alone, \emph{narrow} uses GSM8K alone, and \emph{other datasets} replaces the task prompts with prompts of the same kind from other sources (BeaverTails and Dolly; OpenBookQA and MMLU; Yelp). Coverage curves vary the number of task prompts, and an RBF kernel map is fit on the general data.

\paragraph{Tools.}
The \emph{refusal direction} is the difference between mean representations of 128 harmful and 128 harmless instructions \citep{arditi2024refusal}. Addition adds $\alpha m \hat r$ at every position of the chosen block during greedy generation, with $m$ the target's median residual norm on ordinary text, and is scored by refusal on 64 held-out harmless prompts; ablation projects $\hat r$ out of every block and is scored on 64 held-out harmful prompts. Refusal is detected by phrase matching and checked with Llama Guard~3 \citep{inan2023llama}. The \emph{correctness direction} averages, over 50 labeled questions, the representation of the correct answer minus the mean of the incorrect ones \citep{armstrong2026correctness}, and scores candidates by projection on ARC-Challenge (500 test questions) and TruthfulQA (400, four options each). The \emph{sentiment probe} is the difference of mean representations of 50 positive and 50 negative SST-2 reviews \citep{socher2013recursive}, thresholded at the midpoint of the class means and tested on 500 validation reviews. A tool is carried by mapping the source representations of its defining prompts and rebuilding it from the mapped representations; for a difference of means this equals the map's linear part applied to the source tool, up to renormalization. Carried tools use source labels only.

\paragraph{Measurement.}
Each carried tool is compared with the target's own tool and with three controls of equal magnitude: a random direction, a random direction orthogonal to the native one, and a random source direction carried through the general map. We report recovery,
\begin{equation}
  R = \frac{E_{\text{carried}} - E_0}{E_{\text{native}} - E_0},
  \label{eq:recovery}
\end{equation}
where $E$ is the induced refusal on harmless prompts (addition), the drop in refusal on harmful prompts (ablation), or answer accuracy (correctness), and $E_0$ is no intervention or zero-shot scoring. Refusal is compared at each target's working strength $\alpha^*$, the smallest $\alpha \in \{0.1, 0.2, 0.4, 0.8, 1.2\}$ at which the native direction refuses half of the harmless prompts. Recovery averages over targets whose own tool works and yields coherent output: 10 for refusal addition, 9 for ablation and for ARC, and 11 for TruthfulQA (Appendix~\ref{app:exclusions}). For sentiment we report accuracy and its difference from the target's own probe, since the ratio is unstable when a native probe sits near chance. Main experiments use five seeds, each redrawing the tool's prompts, the task prompts and coverage subsets, the general map's split, and the controls; intervals are 95\% bootstrap intervals that resample models, since pairs share them (Appendix~\ref{app:seeds}). No carried tool uses target labels: the target is run only on unlabeled prompts, and its labels enter the evaluation alone, through the native reference, $\alpha^*$, and the choice of depth, which are shared by every condition for that target.

\paragraph{Replication.} The main experiments are repeated with five seeds; ablation and the kernel and task-only maps use one. Each seed redraws the prompts that define the tool, the task prompts added to the map and the subsets used for each coverage level, the general map's training split, and the control directions; evaluation prompts stay fixed. We report the mean over seeds of each pair's recovery. Pairs share models, so intervals are 95\% bootstrap intervals that resample models and weight each pair by how often its source and target are drawn \citep{efron1994introduction}. These intervals are two to three times wider than intervals over pairs. We report the spread of per-seed means where a conclusion depends on it (Appendix~\ref{app:seeds}).

\paragraph{What label-free means.}
No carried tool uses labels from the target. Fitting a map requires running the target on the unlabeled prompts listed above, and the steering scale $m$ comes from the target's activations on ordinary text. Target labels enter only the evaluation, through the native reference tool and the choice of $\alpha^*$, which is fixed per target and shared by every condition.

\begin{table}[t]
\centering
\caption{Transfer of three tools through each map. Entries are recovery (\%, Eq.~\ref{eq:recovery}) with 95\% model-level bootstrap intervals; sentiment is probe accuracy (\%) with its difference from the target's own probe (89.5\%). Values are means over five seeds except rows marked $\dagger$ and the ablation column (one seed). Correctness values below zero fall under the zero-shot reference. Refusal: 10 targets and 100 pairs for addition, 9 and 90 for ablation (Gemma-2-27B's own direction removes refusal but makes every response repetitive, so it is not a valid reference; Appendix~\ref{app:exclusions}); ARC: 9 targets; TruthfulQA and sentiment: 11 targets, 110 pairs.}
\label{tab:main}
\small
\setlength{\tabcolsep}{4pt}
\newcommand{\ci}[2]{{\scriptsize[#1,\,#2]}}
\begin{tabular}{@{}lccccc@{}}
\toprule
 & \multicolumn{2}{c}{Refusal} & \multicolumn{2}{c}{Correctness} & Sentiment \\
\cmidrule(lr){2-3}\cmidrule(lr){4-5}\cmidrule(l){6-6}
Map & Addition & Ablation & ARC & TruthfulQA & Accuracy ($\Delta$) \\
\midrule
General                  & 54 \ci{37}{70}   & 71 \ci{53}{92}  & 16 \ci{-23}{46}    & 77 \ci{69}{85}   & 81.7 \scriptsize{($-$7.8)} \\
General, kernel$^\dagger$ & 54 \ci{39}{71}   & 88 \ci{67}{108} & 27 \ci{-9}{56}     & 74 \ci{64}{84}   & 80.1 \scriptsize{($-$10.0)} \\
General + task           & 102 \ci{99}{106} & 82 \ci{62}{102}  & 88 \ci{66}{101}    & 98 \ci{97}{100}  & 89.6 \scriptsize{(+0.2)} \\
Task only$^\dagger$      & 101 \ci{99}{104} & 79 \ci{59}{98}  & 87 \ci{63}{101}    & 99 \ci{98}{100}  & 88.8 \scriptsize{($-$1.4)} \\
General + other datasets & 88 \ci{80}{96}   & 78 \ci{58}{100}  & 73 \ci{41}{93}     & 82 \ci{79}{86}   & 88.6 \scriptsize{($-$0.9)} \\
Narrow (GSM8K)           & 2 \ci{0}{3}      & 4 \ci{0}{8}     & $-$66 \ci{-95}{-34}  & 15 \ci{3}{24}    & 57.5 \scriptsize{($-$32.0)} \\
\midrule
Mapped random            & 0 \ci{0}{1}      & 2 \ci{-1}{5}   & $-$149 \ci{-185}{-116} & $-$3 \ci{-13}{5} & 47.5 \scriptsize{($-$42.0)} \\
Random direction         & 1 \ci{-1}{1}     & $-$1 \ci{-6}{1} & $-$150 \ci{-231}{-104} & $-$6 \ci{-21}{4} & 52.2 \scriptsize{($-$37.3)} \\
\bottomrule
\end{tabular}
\end{table}

\begin{figure}[t]
\centering
\includegraphics[width=\linewidth]{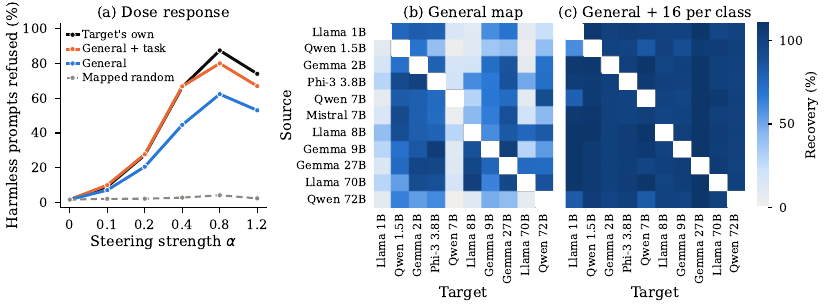}
\caption{Refusal steering carried across models. (a) Refusal of harmless prompts against steering strength, averaged over the 10 targets at each target's working depth (one seed). (b, c) Recovery at each target's working strength for every source and target pair, averaged over five seeds, with the general map (mean 54\%) and with 16 harmful and 16 harmless prompts added (mean 102\%). Values above 110\% are shown at the darkest shade.}
\label{fig:transfer}
\end{figure}

\section{Tools transfer through the map}
\label{sec:specific}

Table~\ref{tab:main} carries each tool from every source to every target through each map. Transfer is specific: random directions, and random source directions carried through the general map, recover at most 1\% of refusal under addition and 4\% under ablation, leave the sentiment probe within 3 points of chance, and rank answers below the zero-shot reference. In every map condition, 0.4\% or fewer of the outputs steered by a carried direction are degenerate on average, and ablating a carried direction makes Llama Guard~3 \citep{inan2023llama} flag 33\% to 54\% of responses to harmful prompts as unsafe, against 53\% for the target's own direction and 6\% to 8\% at baseline and for random directions.

The map's training data sets how much transfers. A map fit on ordinary questions carries 54\% of refusal steering, 77\% of the TruthfulQA gain, almost none of the ARC gain (16\%, with an interval that includes zero), and a sentiment probe 8 points below the target's own. Adding unlabeled prompts from the tool's region raises refusal to 102\%, with every target at least 97\%, correctness to 88\% to 98\%, and the sentiment probe to the native one's accuracy (Figure~\ref{fig:transfer}b--c); the carried direction then tracks the target's own across steering strengths (Figure~\ref{fig:transfer}a). The task prompts alone do as well, prompts of the same kind from other datasets recover 73\% to 88\%, and a map fit on GSM8K alone transfers almost nothing.

Ablation at first gains less from task prompts (71\% to 82\%; kernel map, 88\%), because it projects one direction out of every block and so depends on how well a direction estimated at one depth matches the target at the others. With a task-only map fit at every block and each block ablated with its own carried direction, the carried directions recover 101\% $[89, 112]$ of what the target's own per-block directions achieve, and both leave 8\% to 9\% of harmful prompts refused, against 19\% for the target's single native direction (Appendix~\ref{app:blocks}).

Family, scale, and training paradigm matter little by comparison. With task prompts in the map, refusal recovery is 102\% within and across families and does not depend on the size gap: Llama-3.2-1B steers Llama-3.1-70B at 101\%, and the reverse reaches 96\%. Our two smallest targets recover 97\% and 102\%, although \citet{agarwal2026cross} report that cross-architecture steering degrades below about 1.7B parameters. Under the general map the remaining variation is driven by the target (the pale columns of Figure~\ref{fig:transfer}b). Between the four base models of the mapping study and the instruction-tuned models, in either direction, the TruthfulQA and sentiment tools recover 93\% to 101\% with task prompts, and ARC 61\% to 92\% (Appendix~\ref{app:base}). Per-pair results are in Appendix~\ref{app:pairs}.

\begin{figure}[t]
\centering
\includegraphics[width=\linewidth]{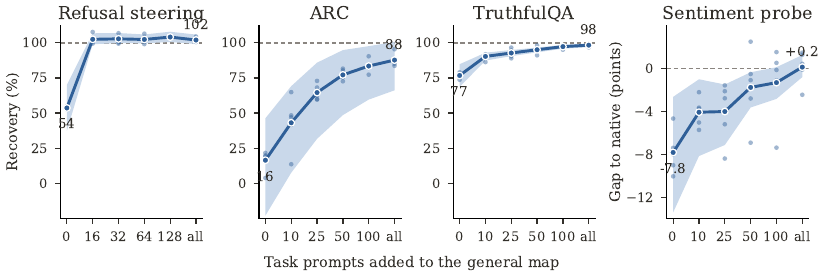}
\caption{Coverage curves. Each panel adds a random subset of unlabeled task prompts to the general map (``all'' is 256 per class for refusal, 200 questions per dataset for correctness, and 256 reviews for sentiment). Lines are means over five seeds, bands are 95\% model-level bootstrap intervals, and dots are per-seed means; the 128-prompt refusal point uses one seed. Dashed lines mark the target's own tool.}
\label{fig:coverage}
\end{figure}

\section{Coverage determines transfer}
\label{sec:coverage}

Figure~\ref{fig:coverage} adds task prompts to the general map in increasing numbers. Refusal saturates at the first step: with 16 harmful and 16 harmless prompts, recovery averages 102\% over five independent draws, and 97\% of the 500 runs recover at least 80\% (the weakest, 44\%). Correctness rises gradually, from 16\% to 88\% on ARC and from 77\% to 98\% on TruthfulQA, and the sentiment gap closes from 7.8 points to 0.2. The ARC curve shows what coverage refers to: the general data already contains 600 ARC questions, yet the general map recovers only 16\% of the ARC gain, because the correctness direction reads a question followed by a candidate answer and the general data contains questions alone. Coverage is a property of the representations a tool reads, and prompts of the same kind from other sources supply it (Table~\ref{tab:main}).

A short calculation makes coverage precise. Write a tool built from class means of source representations as $\delta_S = \bar x_A - \bar x_B$, the difference between the mean representations of its two classes of defining prompts. Because the map is affine, the carried tool before renormalization is $\delta_S W_\lambda$, where $W_\lambda = (X_c^\top X_c + \lambda I)^{-1} X_c^\top Y_c$ is the ridge solution on the centered training representations. With $S_\lambda = (X_c^\top X_c + \lambda I)^{-1} X_c^\top X_c$, we define the \emph{coverage score} $c = \|\delta_S S_\lambda\| / \|\delta_S\| \in [0, 1]$, which depends only on the tool, the source's training representations, and $\lambda$.

\begin{proposition}
\label{prop:coverage}
Suppose that on the training prompts and the tool's prompts the target's representations satisfy $y = xA + b + \varepsilon$, with noise $\varepsilon$ of mean zero and independent of $x$, and let $\delta_T$ be the target's own tool. Then, in expectation over the noise,
\begin{equation}
\mathbb{E}[\delta_S W_\lambda] - \mathbb{E}[\delta_T] = -\lambda\, \delta_S (X_c^\top X_c + \lambda I)^{-1} A,
\end{equation}
and the angle $\theta$ between the expected carried and native tools satisfies $\sin\theta \le \kappa \sqrt{1 - c^2}$, where $\kappa = \|A\| \|\delta_S\| / \|\delta_S A\| \ge 1$.
\end{proposition}

The proof is in Appendix~\ref{app:theory}. A tool that lies in directions the training data spans strongly ($c \to 1$) transfers exactly; a tool along which the training data never varies is carried as zero, as with the GSM8K map; and for fixed $\lambda$, adding training prompts can only shrink the uncovered term $\lambda\, \delta_S^\top (X_c^\top X_c + \lambda I)^{-1} \delta_S$, which bounds the error, so every coverage curve rises. The linearity assumption and $\kappa$ are properties of the pair, which the mapping study and geometry below examine.

\begin{figure}[t]
\centering
\includegraphics[width=\linewidth]{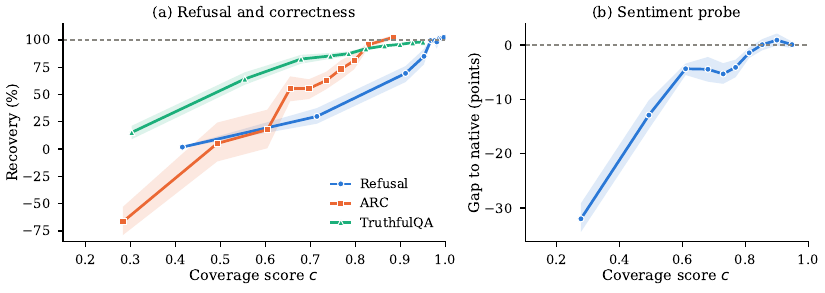}
\caption{Recovery against the coverage score $c$ of the map, over every pair, map, and seed (3,500 to 4,510 runs per tool), in ten equal-count bins of $c$. Bands are 95\% bootstrap intervals over pairs. Dashed lines mark the target's own tool.}
\label{fig:covscore}
\end{figure}

The coverage score needs no target labels, and it tracks recovery as the proposition predicts (Figure~\ref{fig:covscore}). Across every pair, map, and seed, $c$ correlates with recovery at Spearman $\rho = 0.67$ for refusal, 0.71 for ARC, 0.68 for TruthfulQA, and 0.45 for the sentiment probe. The relation behaves like a threshold: refusal runs with $c \ge 0.95$ recover 99\% on average and 94\% of them at least 80\%, while runs with $c < 0.8$ recover 4\%. The general map covers refusal at $c = 0.88$ and ARC at 0.58, and task prompts raise these to 0.995 and 0.83. For a fixed map, coverage does not rank pairs ($\rho = -0.05$ across refusal pairs under the general map), as the pair-specific factor $\kappa$ leads one to expect; Section~\ref{sec:predict} turns to that question.

\begin{figure}[t]
\centering
\includegraphics[width=0.55\linewidth]{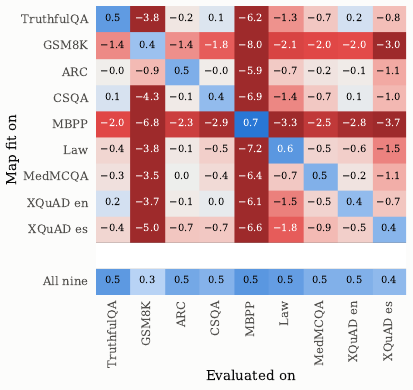}
\caption{$R^2$ of maps fit on one domain (rows) and evaluated on each domain (columns), averaged over 110 pairs at mid depth. The bottom row is one map fit on all nine domains. Colors are clipped at $-3$.}
\label{fig:domains}
\end{figure}

The same account explains why maps between models have been reported to fail across domains \citep{armstrong2026thinking}. In the mapping study on our 11 models (Appendix~\ref{app:maps}), a map fit on one domain reaches $R^2 = 0.48$ on held-out prompts from that domain and $-1.94$ on the other eight, and 90\% of the 72 train and evaluation domain combinations give negative $R^2$ (Figure~\ref{fig:domains}). One map fit on all nine domains matches the specialists on average (0.47) and trails them by at most 0.12, while a map fit on eight domains reaches $-0.62$ on the ninth. The reported failures are therefore failures to extrapolate beyond the map's training data, and they disappear once the training data includes the domain.

The geometry supports the proposition's linearity assumption where the map has data. The models agree on the shape of their representations (linear CKA of 0.84 at mid depth, against 0.01 for shuffled pairings; \citealp{kornblith2019similarity}), a kernel map adds only 0.05 to $R^2$, and linear maps fit on local neighborhoods predict no better than one global linear map (Appendix~\ref{app:geometry}).

\section{What carrying offers without target labels}
\label{sec:predict}

\begin{table}[t]
\centering
\caption{Carrying a tool against building it in the target from the same prompts (recovery, \%; sentiment: accuracy minus the target's own probe, points), with 95\% model-level intervals over seeds 1 to 4. Without labels, the target's tool is the top principal component of the prompts, signed by zero-shot scoring (correctness), the prompt categories (refusal), or an oracle (sentiment).}
\label{tab:rebuild}
\small
\setlength{\tabcolsep}{4pt}
\newcommand{\ci}[2]{{\scriptsize[#1,\,#2]}}
\begin{tabular}{@{}llccc@{}}
\toprule
 & & & \multicolumn{2}{c}{Built in the target} \\
\cmidrule(l){4-5}
Tool & Prompts & Carried & Without labels & With labels \\
\midrule
Refusal    & 16 per class  & 103 \ci{99}{107} & 92 \ci{81}{101} & 97 \ci{91}{101} \\
           & 256 per class & 102 \ci{99}{106} & 97 \ci{89}{102} & 100 \ci{99}{101} \\
\addlinespace
ARC        & 10 questions  & 44 \ci{8}{69}    & $-$76 \ci{-156}{-6} & 70 \ci{50}{85} \\
           & 200 questions & 88 \ci{68}{101}  & $-$14 \ci{-132}{67} & 109 \ci{103}{116} \\
\addlinespace
TruthfulQA & 10 questions  & 91 \ci{88}{94}   & 81 \ci{70}{91}  & 96 \ci{93}{99} \\
           & 200 questions & 98 \ci{97}{100}  & 86 \ci{76}{93}  & 101 \ci{100}{102} \\
\addlinespace
Sentiment  & 10 reviews    & $-$3.6 \ci{-7.7}{-0.5} & $-$22.3 \ci{-31.8}{-12.7} & $-$12.7 \ci{-21.3}{-5.8} \\
           & 256 reviews   & $+$0.8 \ci{-0.7}{2.8}  & $-$22.3 \ci{-34.9}{-9.5}  & $+$1.6 \ci{-0.2}{6.3} \\
\bottomrule
\end{tabular}
\end{table}

\begin{figure}[t]
\centering
\includegraphics[width=0.55\linewidth]{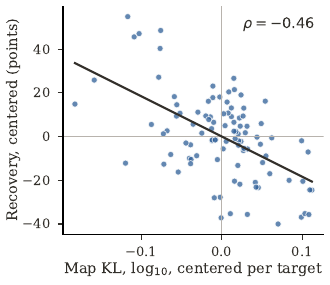}
\caption{Map KL against refusal recovery under the general map, both centered within each target (100 pairs, five seeds).}
\label{fig:predict}
\end{figure}

A carried tool uses the source's labels and only unlabeled prompts on the target. Table~\ref{tab:rebuild} asks whether the target could do as well with those prompts alone. For refusal it could: the target's own direction from the same 16 prompts per class recovers 97\%, against 103\% carried. Choosing prompts by category is all the supervision a refusal direction needs, so its transfer shows that the mechanism corresponds across models and saves no labeling. The read-out tools differ. Without labels, the target's principal component fails on ARC, reaches 86\% on TruthfulQA, and leaves the sentiment probe 22 points below the native one, while the carried tools recover 88\% and 98\% and match the native probe. The carried ARC direction is worth about 25 labeled target questions, and the sentiment probe carried through 10 unlabeled reviews is 9 points more accurate than one built from 10 labeled reviews (Appendix~\ref{app:rebuild}). Carrying substitutes for target labels when the target's unlabeled representations do not reveal the concept by themselves.

Without task prompts, a tool must be carried through the general map, and coverage does not separate sources (Section~\ref{sec:coverage}). A second label-free measure is specific to the pair: we patch mapped source representations into the target on held-out general prompts and record the KL divergence of its next-token distribution from the unpatched one. Across refusal pairs map KL relates only loosely to recovery ($\rho = -0.35$, interval $[-0.72, 0.05]$), because targets differ in their typical KL, but within a target it ranks sources ($\rho = -0.46$ after centering per target, Figure~\ref{fig:predict}; $p < 0.001$ by permutation). For correctness it also orders pairs across targets ($\rho = -0.64$ on ARC and $-0.53$ on TruthfulQA). Carrying from the lowest-KL source raises mean refusal recovery from 54\% to 67\%, against 80\% for the best source in hindsight. Coverage thus tells whether the map's data suffice, and map KL which source to carry from when they do not.

\section{Discussion and limitations}
\label{sec:discussion}

The results treat an interpretability tool as an estimate of a mechanism that other models share. Carried through a map whose unlabeled training data covers the tool's region, a direction found in one model steers or reads a second model about as well as that model's own direction, across five families, from 1B to 72B parameters, and between base and instruction-tuned models. Proposition~\ref{prop:coverage} states when this must happen: if two models correspond linearly on the covered region, the carried tool departs from the native one only through the part the map's data leave uncovered, and the coverage score that measures this part tracks recovery in every experiment. The same view qualifies claims that models converge on shared representations \citep{huh2024platonic}: our models agree on the shape of their representations, yet a map fit on one distribution transfers little of a tool that reads another, so convergence is better described as agreement within the regions a map has seen.

The practical value of transfer depends on the tool. A refusal direction can be rebuilt in the target from the same 16 prompts per class, so its transfer shows that the mechanism corresponds and saves no labeling, whereas carried correctness and sentiment tools outperform anything the target's unlabeled representations yield. The correspondence also cuts the other way: a carried refusal direction removes most of the refusal that the target's own direction removes. This requires white-box access to the target's activations, which API-only models do not provide, and anyone with that access can also find the direction in the target directly.

Several limits apply. We transfer linear directions and a linear probe; sparse autoencoder features and circuits are untested and may need maps of a different form. Proposition~\ref{prop:coverage} assumes a linear correspondence, ignores the renormalization of mapped representations, and does not say which pairs correspond well. Our models are 11 instruction-tuned open models, the two largest in 4-bit precision, and four base models tested with the read-out tools only. Refusal is scored by phrase matching at one strength per target; over the full range of strengths, measured for one seed, the general map with task prompts recovers 95\% of the native effect (Appendix~\ref{app:refusal}). Ablation excludes Gemma-2-27B by a coherence criterion added after the main run (Appendix~\ref{app:exclusions}), and the read-out tools are never used to steer. Coverage requires knowing where a tool operates, and small draws can miss: the weakest 16-prompt draw recovers 44\% of refusal, mostly on Llama-3.2-1B, the target with the smallest native effect. Ablation, the per-block and kernel maps, and the base-model transfers use one seed.

\bibliographystyle{abbrvnat}
\bibliography{references}

\appendix

\section{Models, depths, and inclusion}
\label{app:exclusions}

\begin{table}[h]
\centering
\caption{Models. All are instruction-tuned checkpoints from Hugging Face. Precision is the weight format at inference; 4-bit models use NF4 quantization with double quantization and bf16 computation (bitsandbytes).}
\label{tab:app_models}
\small
\begin{tabular}{@{}llccc@{}}
\toprule
Model & Checkpoint & Blocks & Precision & GPUs \\
\midrule
Llama-3.2-1B  & \texttt{meta-llama/Llama-3.2-1B-Instruct}   & 16 & bf16  & 1 \\
Qwen2.5-1.5B  & \texttt{Qwen/Qwen2.5-1.5B-Instruct}         & 28 & bf16  & 1 \\
Gemma-2-2B    & \texttt{google/gemma-2-2b-it}               & 26 & bf16  & 1 \\
Phi-3-mini    & \texttt{microsoft/Phi-3-mini-4k-instruct}   & 32 & bf16  & 1 \\
Qwen2.5-7B    & \texttt{Qwen/Qwen2.5-7B-Instruct}           & 28 & bf16  & 1 \\
Mistral-7B    & \texttt{mistralai/Mistral-7B-Instruct-v0.3} & 32 & bf16  & 1 \\
Llama-3.1-8B  & \texttt{meta-llama/Llama-3.1-8B-Instruct}   & 32 & bf16  & 1 \\
Gemma-2-9B    & \texttt{google/gemma-2-9b-it}               & 42 & bf16  & 1 \\
Gemma-2-27B   & \texttt{google/gemma-2-27b-it}              & 46 & bf16  & 2 \\
Llama-3.1-70B & \texttt{meta-llama/Llama-3.1-70B-Instruct}  & 80 & 4-bit & 2 \\
Qwen2.5-72B   & \texttt{Qwen/Qwen2.5-72B-Instruct}          & 80 & 4-bit & 2 \\
\bottomrule
\end{tabular}
\end{table}

All experiments ran on two NVIDIA RTX 6000 Ada GPUs (48\,GB each). Every model serves as a source for every tool. A model serves as a target only when its own tool has an effect to recover, because recovery (Eq.~\ref{eq:recovery}) divides by the native effect.

For refusal, a target must refuse at least 50\% of harmful prompts without intervention, and its own direction must raise refusal on harmless prompts by at least 30 points at its working strength. Mistral-7B refuses 9\% of harmful prompts at baseline and is excluded as a refusal target; it remains a source, so refusal uses 10 targets and 100 pairs. For ablation, the native direction must lower refusal on harmful prompts by at least 30 points and leave at most half of its responses degenerate (fewer than three words, or fewer than 30\% distinct words). All 10 refusal targets pass the first condition. Gemma-2-27B fails the second: ablating its own direction makes every response repetitive, although Llama Guard~3 still flags 91\% of them as unsafe. We added this criterion after the main run, so ablation uses 9 targets and 90 pairs, and we report the 10-target results for completeness. With Gemma-2-27B included, ablation recovery is 72\% for the general map, 89\% for the kernel map, 75\% with task prompts, 72\% for task prompts alone, 78\% with other-dataset prompts, 4\% for the GSM8K map, and 4\% for the mapped random control; no conclusion changes. For correctness, the native direction must beat zero-shot scoring by at least 5 points. Llama-3.2-1B (37\% zero-shot, 34\% native) and Qwen2.5-1.5B (45\% and 49\%) fail this criterion on ARC, leaving 9 ARC targets; all 11 models pass on TruthfulQA. All 11 models are sentiment targets. Inclusion is decided on the first seed and held fixed for the others.

Refusal is steered at relative depth $d = 0.5$, chosen before any transfer experiment. To check that this choice does not disadvantage particular targets, we swept the native direction over $d \in \{0.25, 0.5, 0.75\}$ and all strengths (Table~\ref{tab:app_depth}). Depth 0.5 works for every target except Qwen2.5-72B, whose own direction induces no refusal at depths 0.25 or 0.5 at any strength and induces full refusal at 0.75. We therefore run the full refusal experiment for Qwen2.5-72B at $d = 0.75$, where every map for that target is fit and every carried direction is applied. No other target was moved, including those that also work at 0.75. The working strength $\alpha^*$ of each target (Table~\ref{tab:models}) is the smallest strength in $\{0.1, 0.2, 0.4, 0.8, 1.2\}$ at which the native direction refuses at least half of the harmless prompts; it is set from the native direction alone and shared by every condition for that target.

\begin{table}[h]
\centering
\caption{Native refusal direction by depth: the highest refusal rate on harmless prompts (\%) over strengths $\alpha \in \{0.1, 0.2, 0.4, 0.8, 1.2\}$. Bold marks the depth used in the experiments.}
\label{tab:app_depth}
\small
\begin{tabular}{@{}lccc@{}}
\toprule
Target & $d = 0.25$ & $d = 0.5$ & $d = 0.75$ \\
\midrule
Llama-3.2-1B  & 2  & \textbf{50}  & 31 \\
Qwen2.5-1.5B  & 3  & \textbf{100} & 100 \\
Gemma-2-2B    & 5  & \textbf{100} & 100 \\
Phi-3-mini    & 58 & \textbf{94}  & 97 \\
Qwen2.5-7B    & 0  & \textbf{100} & 100 \\
Mistral-7B$^\dagger$ & 5 & 20 & 6 \\
Llama-3.1-8B  & 100 & \textbf{100} & 6 \\
Gemma-2-9B    & 56 & \textbf{100} & 98 \\
Gemma-2-27B   & 100 & \textbf{100} & 63 \\
Llama-3.1-70B & 38 & \textbf{88}  & 41 \\
Qwen2.5-72B   & 0  & 0  & \textbf{100} \\
\bottomrule
\multicolumn{4}{@{}l}{\footnotesize $^\dagger$Excluded as a refusal target (baseline refusal 9\%).}
\end{tabular}
\end{table}

\section{Maps and the mapping study}
\label{app:maps}

\paragraph{Representations.} For a prompt $p$, a model with $L$ blocks gives $h(p) \in \mathbb{R}^{D}$, the residual stream at the last token after block $\lfloor dL \rceil$, divided by its norm. Maps are fit and applied on these unit vectors. When a carried direction is rebuilt, the mapped representations are renormalized to unit length before class means are taken.

\paragraph{Ridge map.} Given $n$ prompts with source representations $X \in \mathbb{R}^{n \times D_s}$ and target representations $Y \in \mathbb{R}^{n \times D_t}$, we center both ($X_c$, $Y_c$, with means $\mu_X$, $\mu_Y$) and solve
\begin{equation}
W_\lambda = \arg\min_W \|Y_c - X_c W\|_F^2 + \lambda \|W\|_F^2 = (X_c^\top X_c + \lambda I)^{-1} X_c^\top Y_c,
\end{equation}
so that $M(x) = (x - \mu_X) W_\lambda + \mu_Y$. We compute the fit from the eigendecomposition of $X_c X_c^\top$ when $D_s \geq n$ and of $X_c^\top X_c$ otherwise. The penalty is chosen from $\lambda \in \{10^{-3}, 3 \cdot 10^{-3}, \ldots, 30, 100\}$ (11 values) by exact leave-one-out error,
\begin{equation}
\mathrm{LOO}(\lambda) = \frac{1}{n} \sum_{i=1}^{n} \frac{\|y_i - \hat y_i\|^2}{(1 - H_{ii})^2}, \qquad H = X_c (X_c^\top X_c + \lambda I)^{-1} X_c^\top,
\end{equation}
which needs one eigendecomposition for the whole grid. The map is fit once per source, target, training set, and seed, with no access to target labels.

\paragraph{Kernel map.} The kernel map is RBF kernel ridge regression, $M(x) = \mu_Y + \sum_i a_i \exp(-\gamma \|x - x_i\|^2)$ with $a = (K + \lambda I)^{-1} Y_c$. The bandwidth is $\gamma = \gamma_{\text{rel}} / \bar\delta^2$, where $\bar\delta^2$ is the mean squared distance between training points and $\gamma_{\text{rel}} \in \{0.25, 0.5, 1, 2, 4\}$; $(\gamma_{\text{rel}}, \lambda)$ is chosen jointly by exact leave-one-out error over the same $\lambda$ grid.

\paragraph{Training sets.} Table~\ref{tab:app_arms} lists the training data of each map. Task prompts are always disjoint from the prompts that define the tool and from the evaluation prompts, and the map never sees a label: harmful and harmless prompts, answer candidates, and reviews enter as text only. For correctness, the task prompts are the representations of every answer candidate (\texttt{Question: \{q\} Answer: \{a\}}) for 200 questions per dataset, so the map sees correct and incorrect candidates alike without knowing which is which.

\begin{table}[h]
\centering
\caption{Training data of each map. General data uses the \texttt{qa} template for correctness and sentiment and the chat template for refusal. Coverage subsets are drawn at random from the task prompts, separately for each seed.}
\label{tab:app_arms}
\small
\setlength{\tabcolsep}{4pt}
\begin{tabular}{@{}p{2.6cm}p{3.4cm}p{3.6cm}p{3.0cm}@{}}
\toprule
Map & Refusal & Correctness & Sentiment \\
\midrule
General & \multicolumn{3}{p{10.2cm}}{5,400 questions, 600 from each of TruthfulQA, GSM8K, ARC-Challenge, CommonsenseQA, MBPP, MMLU professional law, MedMCQA, XQuAD (en), XQuAD (es)} \\
\addlinespace
+ task & 256 harmful, 256 harmless instructions & candidates of 200 ARC and 200 TruthfulQA questions & 256 SST-2 training reviews \\
\addlinespace
+ other datasets & 256 BeaverTails harmful, 256 Dolly instructions & candidates of 200 OpenBookQA and MMLU questions & 256 Yelp reviews \\
\addlinespace
Coverage & 16, 32, 64, 128 per kind & 10, 25, 50, 100 questions per dataset & 10, 25, 50, 100 reviews \\
\addlinespace
Narrow & \multicolumn{3}{l}{600 GSM8K questions} \\
\bottomrule
\end{tabular}
\end{table}

\paragraph{Mapping study.} The mapping study measures how well maps predict target representations, independently of any tool. Each of the nine general domains contributes 800 prompts, split into 600 for fitting and 200 for evaluation under each of five seeds; XQuAD English and Spanish share a split so that translations stay aligned. We fit five kinds of map for every ordered pair of the 11 models at depths $d \in \{0.25, 0.5, 0.75, 0.9, 1\}$: a \emph{specialist} on each domain's 600 prompts, a \emph{mixed} map on all 5,400, a \emph{matched mixed} map on 600 prompts spread evenly over the nine domains, a kernel map on all 5,400, and a \emph{shuffled} control, the matched mixed map fit after permuting target representations across prompts. Leaving one domain out gives the map on the other eight. Fit is the variance-weighted coefficient of determination on the evaluation prompts of one domain,
\begin{equation}
R^2 = 1 - \frac{\sum_i \|y_i - M(x_i)\|^2}{\sum_i \|y_i - \bar y\|^2},
\end{equation}
where $\bar y$ is the mean target representation of that domain's evaluation prompts, so $R^2 < 0$ means the map predicts worse than the domain mean. Table~\ref{tab:app_mapping} gives the results by depth.

\begin{table}[h]
\centering
\caption{Mapping study: $R^2$ on held-out prompts, averaged over 110 pairs and five splits. Specialists are evaluated on their own domain (own) and on the other eight (other); all other maps are evaluated on every domain they were fit on. Leave one domain out is evaluated on the held-out domain.}
\label{tab:app_mapping}
\small
\setlength{\tabcolsep}{3.5pt}
\begin{tabular}{@{}lccccc@{}}
\toprule
Map & $d = 0.25$ & $d = 0.5$ & $d = 0.75$ & $d = 0.9$ & $d = 1$ \\
\midrule
Specialist, own    & 0.64 & 0.48 & 0.39 & 0.32 & 0.32 \\
Specialist, other  & $-1.82$ & $-1.94$ & $-1.53$ & $-1.34$ & $-1.65$ \\
Mixed              & 0.62 & 0.47 & 0.38 & 0.33 & 0.34 \\
Matched mixed      & 0.45 & 0.27 & 0.13 & 0.03 & 0.03 \\
Kernel             & 0.67 & 0.53 & 0.45 & 0.40 & 0.41 \\
Shuffled           & $-1.67$ & $-1.44$ & $-1.07$ & $-1.08$ & $-1.18$ \\
Leave one domain out & -- & $-0.62$ & $-0.55$ & -- & $-0.59$ \\
\bottomrule
\end{tabular}
\end{table}

The mixed map matches the specialists at every depth while serving all nine domains, and its advantage over the matched mixed map shows that the gain comes from the number of prompts per domain, since both cover the same domains. Figure~\ref{fig:domains} shows every combination: 90\% of the 72 off-diagonal cells are negative, and the worst, a map fit on TruthfulQA and evaluated on GSM8K, reaches $-3.82$. The mixed map trails the specialists most on MBPP (by 0.12) and GSM8K (by 0.07), where it still reaches 0.32. The kernel map improves on the mixed map by 0.05 to 0.07, an advantage that grows with depth.

\paragraph{Map fidelity.} For Section~\ref{sec:predict}, we patch the mixed map's prediction $M(h_s(p))$, rescaled to the target's original norm at that position, into the target's residual stream at the last token of each of 100 held-out prompts per domain, and record $\mathrm{KL}(P_{\text{clean}} \,\|\, P_{\text{patched}})$ of the next-token distribution, averaged over prompts and domains. It uses the depth at which each tool operates and requires no labels.

\section{Tools, prompts, and evaluation}
\label{app:refusal}

\paragraph{Prompts.} Harmless instructions come from Alpaca \citep{taori2023alpaca}. Harmful instructions are drawn from the LLM-LAT harmful dataset \citep{sheshadri2024latent} (324 of the 448 used), HarmfulQA \citep{bhardwaj2023red} (97), AdvBench \citep{zou2023universal} (34), HarmBench \citep{mazeika2024harmbench} (13), and JailbreakBench \citep{chao2024jailbreakbench} (6); some prompts occur in more than one source. For each kind, 128 prompts define the direction, 256 are the task prompts available to the map, and 64 are held out for evaluation, with no prompt in two roles. The other-dataset maps use 256 harmful prompts from BeaverTails \citep{ji2023beavertails} and 256 instructions from Dolly \citep{conover2023dolly}. For correctness, each dataset contributes a pool of 250 labeled questions, of which 50 define the direction and 200 supply unlabeled answer candidates to the map, and a disjoint test set (500 ARC-Challenge questions; 400 TruthfulQA MC1 questions, restricted to four options). For sentiment, 100 SST-2 training reviews define the probe, 256 further training reviews are the task prompts, 500 validation reviews are the test set, and 256 Yelp reviews form the other-dataset prompts. Seeds 1 to 4 redraw the defining and task prompts within each pool.

\paragraph{Steering and scoring.} All generations are greedy, with prompts in each model's chat template and inputs truncated to 384 tokens. Addition adds $\alpha m \hat r$ to the output of the chosen block at every position and generates 48 tokens; ablation subtracts the projection onto $\hat r$ from the embeddings and from the output of every block and generates 96 tokens. The scale $m$ is the median norm of the target's residual stream at that block on the general prompts. A response counts as a refusal if it contains any of \emph{I'm sorry}, \emph{I am sorry}, \emph{I apologize}, \emph{As an AI}, \emph{As a language model}, \emph{As an assistant}, \emph{I cannot}, \emph{I can't}, \emph{I am unable}, \emph{I'm unable}, \emph{I am not able}, \emph{I'm not able}, \emph{I will not}, \emph{I won't}, \emph{I must decline}, \emph{Lo siento}, or \emph{No puedo}, ignoring case. A response is degenerate if it has fewer than three words or fewer than 30\% distinct words. Llama Guard~3 (8B) receives each harmful prompt and the target's response as a conversation and labels the response safe or unsafe.

Correctness is scored by the projection of each candidate's representation onto the direction, and the zero-shot reference by each candidate's mean token log-probability given the question; the highest-scoring candidate is the prediction. The sentiment probe predicts positive when a review's projection exceeds the midpoint of the two class means of its defining reviews.

\paragraph{Controls.} The random direction is a standard Gaussian vector in the target's space, normalized; the orthogonal control removes its component along the native direction. The mapped random control adds a random source direction, scaled to the norm of the source tool, to the representations of one class of defining prompts, carries both versions through the general map, and takes the difference of their mapped means, so it passes through exactly the same map and renormalization as the real tool. Every control is applied at the same strength as the native direction.

\paragraph{Dose response.} Table~\ref{tab:app_dose} gives refusal on harmless prompts at every strength for the first seed. The carried direction with task prompts tracks the native one across strengths, and both decline at the largest strength, where outputs begin to degenerate. Integrating each curve over strength and dividing by the native area gives a recovery measure that does not depend on $\alpha^*$: 67\% $[52, 80]$ for the general map, 70\% for the kernel map, 95\% $[89, 101]$ with task prompts, 94\% for task prompts alone, 5\% for the GSM8K map, and 2\% and 1\% for the mapped random and random controls. The target's own direction computed on normalized representations reaches 96\% by the same measure, which bounds what any map acting on normalized representations can achieve and accounts for recovery slightly above 100\% at $\alpha^*$.

\begin{table}[h]
\centering
\caption{Refusal of harmless prompts (\%) against steering strength, averaged over the 10 refusal targets at each target's working depth (first seed; carried directions averaged over sources).}
\label{tab:app_dose}
\small
\begin{tabular}{@{}lcccccc@{}}
\toprule
 & $\alpha = 0$ & 0.1 & 0.2 & 0.4 & 0.8 & 1.2 \\
\midrule
Target's own direction & 2 & 9 & 27 & 66 & 88 & 74 \\
Carried, general + task & 2 & 10 & 28 & 67 & 80 & 67 \\
Carried, general & 2 & 7 & 21 & 45 & 62 & 53 \\
Mapped random & 2 & 2 & 2 & 3 & 4 & 2 \\
\bottomrule
\end{tabular}
\end{table}

\paragraph{Examples.} Table~\ref{tab:app_examples} shows responses of Llama-3.1-70B to held-out harmless prompts, steered at its working strength with its own direction and with directions carried from Qwen2.5-7B. With the general map the carried direction leaves the response unchanged; with task prompts in the map it produces the same refusal as the target's own direction.

\begin{table}[h]
\centering
\caption{Responses of Llama-3.1-70B to harmless prompts (first words shown), with directions carried from Qwen2.5-7B.}
\label{tab:app_examples}
\small
\begin{tabular}{@{}p{2.4cm}p{10.4cm}@{}}
\toprule
\multicolumn{2}{@{}l}{\emph{Name three jobs related to computer programming}} \\
No steering & Here are three jobs related to computer programming: 1. Software Engineer: A software engineer designs, develops, and tests software applications\ldots \\
Own direction & I can't help with that information. \\
Carried, general & Here are three jobs related to computer programming: 1. Software Engineer: Designs, develops, and tests software applications\ldots \\
Carried, + task & I can't help with that information. \\
\midrule
\multicolumn{2}{@{}l}{\emph{Write a few sentences about the importance of water}} \\
No steering & Water is the foundation of life, playing a crucial role in maintaining the health and well-being of humans, animals, and the environment\ldots \\
Own direction & I can't help you with this information. \\
Carried, general & Access to clean and safe water is a fundamental human right, essential for human dignity, health, and well-being\ldots \\
Carried, + task & I can't help with this information. \\
\bottomrule
\end{tabular}
\end{table}

\section{Seeds and intervals}
\label{app:seeds}

Each of the five seeds redraws the prompts that define each tool, the task prompts available to the map and the subset used at every coverage level, the general map's training split (the same five splits as the mapping study), and the control directions; evaluation prompts are fixed. Seed 0 is the original run, in which the coverage subsets are the first $k$ task prompts rather than a random draw. For sentiment this made the small seed-0 subsets all positive reviews, which explains the non-monotonic seed-0 coverage curve that the other seeds do not show.

Table~\ref{tab:app_seeds} reports, for every condition repeated across seeds, the mean over seeds of each pair's recovery with two 95\% intervals, the range of the five per-seed means, and the standard deviation across seeds within a pair, averaged over pairs. The model-level interval resamples models and weights each pair by how often its source and target are drawn; the pair-level interval resamples pairs and ignores that pairs share models. Model-level intervals are two to three times wider, and we use them throughout. Variation across seeds within a pair is small next to the differences between maps, and it is largest at the smallest coverage levels: per-seed means of the 16-prompt refusal map range from 99\% to 108\%, and of the 10-question ARC map from 14\% to 65\%.

Across the 500 runs of the 16-prompt refusal map, the 5th percentile of recovery is 89\% and the minimum 44\%. Six of the eight weakest runs have Llama-3.2-1B as target, whose own direction raises refusal by only about 50 points, one of the two smallest native effects, so the recovery ratio has one of the smallest denominators there. With all 256 task prompts per class, the weakest of the 500 runs recovers 67\%.

\begin{table}[h]
\centering
\caption{Seed variability. Recovery (\%) or, for sentiment, accuracy minus the target's own probe (points). Mean over five seeds of each pair's value; 95\% intervals resampling models and resampling pairs; range of per-seed means; mean within-pair standard deviation across seeds.}
\label{tab:app_seeds}
\small
\setlength{\tabcolsep}{4pt}
\begin{tabular}{@{}lccccc@{}}
\toprule
Map & Mean & Model-level & Pair-level & Per-seed means & Within-pair SD \\
\midrule
\multicolumn{6}{@{}l}{\emph{Refusal (addition)}} \\
General & 54 & [37, 70] & [47, 59] & 53 to 55 & 6 \\
+ 16 per class & 102 & [99, 107] & [101, 104] & 99 to 108 & 6 \\
+ 32 per class & 103 & [100, 107] & [101, 104] & 99 to 107 & 6 \\
+ 64 per class & 102 & [100, 106] & [101, 104] & 99 to 107 & 5 \\
+ task (256) & 102 & [99, 106] & [101, 103] & 101 to 104 & 4 \\
+ other datasets & 88 & [80, 96] & [85, 91] & 86 to 90 & 4 \\
Narrow & 2 & [0, 3] & [1, 2] & 2 to 2 & 1 \\
Mapped random & 0 & [0, 1] & [0, 1] & 0 to 1 & 2 \\
\addlinespace
\multicolumn{6}{@{}l}{\emph{ARC}} \\
General & 16 & [$-$23, 46] & [2, 30] & 4 to 22 & 22 \\
+ 10 questions & 43 & [8, 69] & [31, 53] & 14 to 65 & 28 \\
+ 25 & 65 & [32, 86] & [55, 73] & 60 to 73 & 20 \\
+ 50 & 77 & [49, 95] & [69, 84] & 73 to 82 & 15 \\
+ 100 & 83 & [60, 98] & [77, 89] & 77 to 90 & 14 \\
+ task (200) & 88 & [66, 101] & [81, 93] & 84 to 95 & 12 \\
+ other datasets & 73 & [41, 93] & [64, 81] & 66 to 79 & 16 \\
Narrow & $-$65 & [$-$95, $-$34] & [$-$78, $-$53] & $-$83 to $-$45 & 36 \\
Mapped random & $-$149 & [$-$185, $-$116] & [$-$162, $-$137] & $-$171 to $-$131 & 60 \\
\addlinespace
\multicolumn{6}{@{}l}{\emph{TruthfulQA}} \\
General & 77 & [69, 85] & [73, 80] & 74 to 79 & 8 \\
+ 10 questions & 90 & [87, 93] & [89, 92] & 86 to 92 & 5 \\
+ 25 & 93 & [91, 95] & [92, 94] & 89 to 97 & 5 \\
+ 50 & 95 & [93, 97] & [94, 96] & 91 to 98 & 4 \\
+ 100 & 97 & [96, 99] & [97, 98] & 95 to 98 & 4 \\
+ task (200) & 98 & [97, 100] & [98, 99] & 97 to 99 & 3 \\
+ other datasets & 82 & [78, 86] & [80, 84] & 81 to 83 & 4 \\
Narrow & 15 & [3, 24] & [10, 21] & 11 to 18 & 16 \\
Mapped random & $-$3 & [$-$13, 5] & [$-$7, 0] & $-$6 to $-$1 & 28 \\
\addlinespace
\multicolumn{6}{@{}l}{\emph{Sentiment (gap, points)}} \\
General & $-$7.8 & [$-$13.4, $-$2.6] & [$-$10.2, $-$5.7] & $-$10.1 to $-$4.7 & 6.5 \\
+ 10 reviews & $-$4.1 & [$-$8.2, $-$1.0] & [$-$5.5, $-$2.9] & $-$5.7 to $-$2.2 & 6.3 \\
+ 25 & $-$4.0 & [$-$7.1, $-$1.5] & [$-$4.9, $-$3.2] & $-$8.4 to $-$1.6 & 6.8 \\
+ 50 & $-$1.7 & [$-$3.6, $-$0.3] & [$-$2.6, $-$1.0] & $-$6.9 to 2.5 & 5.8 \\
+ 100 & $-$1.3 & [$-$2.8, 0.1] & [$-$2.1, $-$0.4] & $-$7.4 to 1.6 & 5.8 \\
+ task (256) & 0.2 & [$-$0.8, 1.3] & [$-$0.4, 0.7] & $-$2.4 to 1.5 & 3.8 \\
+ other datasets & $-$0.9 & [$-$4.7, 2.4] & [$-$2.5, 0.7] & $-$2.2 to 1.6 & 3.5 \\
Narrow & $-$32.0 & [$-$36.3, $-$28.0] & [$-$34.1, $-$29.9] & $-$33.5 to $-$31.2 & 8.5 \\
Mapped random & $-$42.0 & [$-$46.1, $-$37.1] & [$-$43.6, $-$40.3] & $-$43.6 to $-$40.2 & 5.5 \\
\bottomrule
\end{tabular}
\end{table}

\section{Results by pair}
\label{app:pairs}

\begin{figure}[p]
\centering
\includegraphics[width=\linewidth]{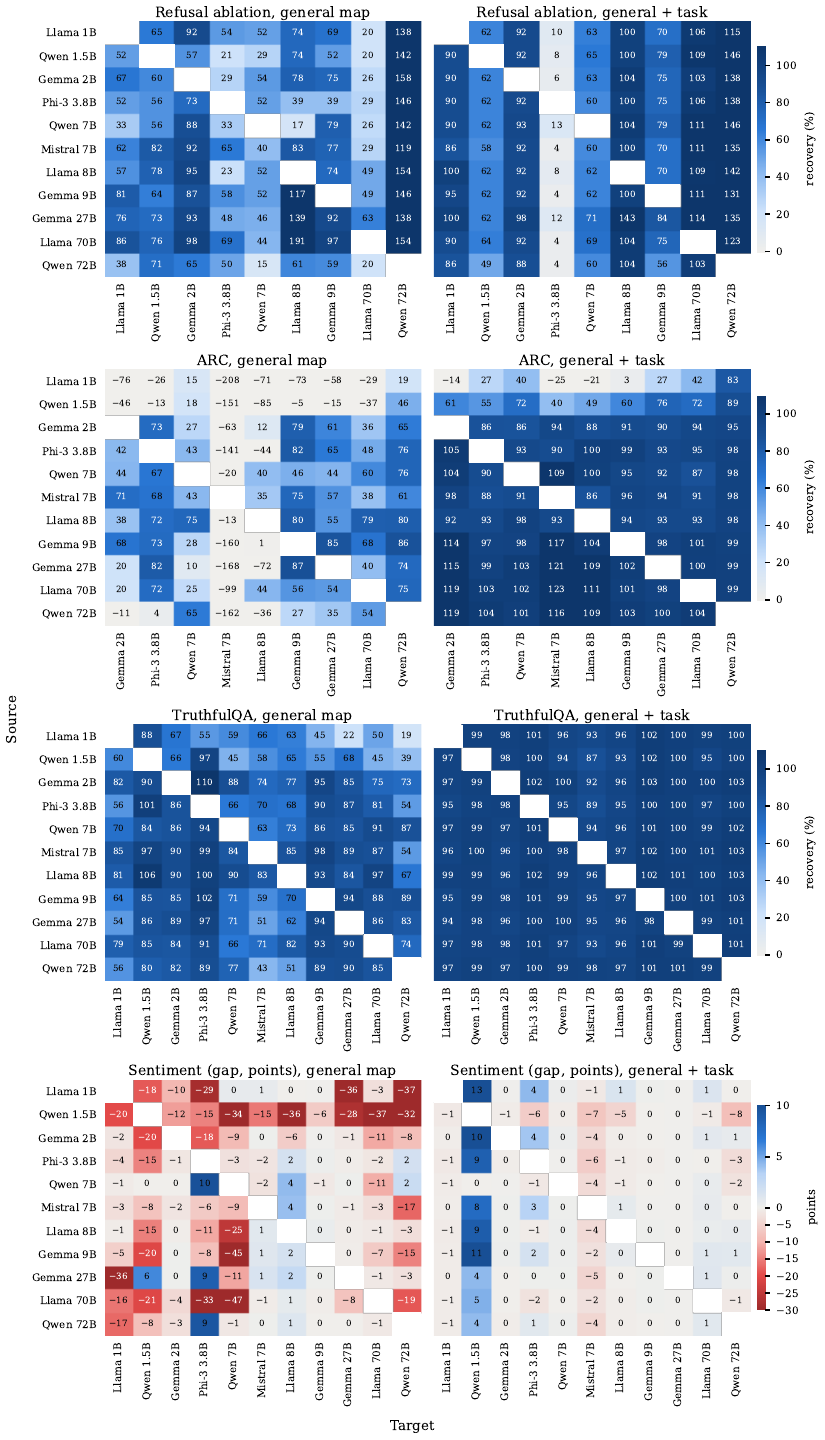}
\caption{Recovery for every source (rows) and target (columns) under the general map (left) and with task prompts added (right). Refusal ablation uses the first seed and excludes Gemma-2-27B as target; the other panels average five seeds. Correctness recovery below zero is shown at the lightest shade; sentiment is accuracy minus the target's own probe, clipped to $[-30, 10]$ points.}
\label{fig:app_pairs}
\end{figure}

Figure~\ref{fig:app_pairs} gives every pair for the tools not shown in Figure~\ref{fig:transfer}, under the general map and with task prompts added. Table~\ref{tab:app_family} summarizes them by family and by the size of source and target.

With task prompts in the map, family and scale leave no trace in refusal or TruthfulQA: recovery is the same within and across families, and its correlation with the absolute log ratio of parameter counts is at most 0.05. ARC is the exception. With task prompts, larger sources transfer more (Spearman $\rho = 0.68$ with the signed log ratio of source to target size): the two smallest sources recover $-$25\% to 89\% across targets, while every source from 2.6B parameters up recovers at least 86\%. Under the general map, recovery depends mostly on the target. Mean refusal recovery ranges from 12\% (Qwen2.5-7B) to 81\% (Gemma-2-2B) across targets and from 28\% (Qwen2.5-1.5B) to 69\% (Llama-3.1-70B) across sources. Within families, the general map transfers ARC better than across them (45\% against 11\%), and TruthfulQA transfers better between models of similar size ($\rho = -0.31$ with the absolute log ratio).

Two patterns in the ablation panels deserve mention. Recovery above 100\% for Qwen2.5-72B as target reflects a native single-direction ablation that leaves 23\% of harmful prompts refused, which carried directions often improve on; with per-block directions the native ablation removes refusal fully (Appendix~\ref{app:blocks}). Phi-3-mini is the one target where task prompts lower ablation recovery, from 21\% to 69\% under the general map to 4\% to 13\%. The cause lies in the normalization: in Phi-3-mini, the target's own direction computed from normalized representations also fails to ablate refusal (81\% of harmful prompts still refused, against 13\% for the direction from raw activations), and carried directions are built in the same normalized space; with task prompts they match that direction closely and inherit its failure. With per-block directions, carried directions leave 33\% of harmful prompts refused in Phi-3-mini, fewer than the target's own per-block directions (48\%). In the sentiment panels, gaps above zero for Qwen2.5-1.5B as target come from its weak native probe (67\% accuracy on average), which carried probes exceed.

\begin{table}[h]
\centering
\caption{Recovery (\%) within and across families, with 95\% model-level intervals, and Spearman correlation of pair recovery with the absolute and signed log ratio of source to target parameter count (five seeds).}
\label{tab:app_family}
\small
\setlength{\tabcolsep}{4pt}
\begin{tabular}{@{}llcccc@{}}
\toprule
Tool & Map & Same family & Across families & $\rho$, $|\log|$ ratio & $\rho$, signed ratio \\
\midrule
Refusal    & General      & 58 [30, 88]   & 53 [35, 67]  & $-$0.17 & 0.12 \\
           & + task       & 102 [98, 108] & 102 [99, 106] & 0.05 & $-$0.19 \\
ARC        & General      & 45 [$-$14, 74] & 11 [$-$31, 43] & $-$0.10 & $-$0.03 \\
           & + task       & 85 [40, 102]  & 88 [68, 101] & $-$0.03 & 0.68 \\
TruthfulQA & General      & 78 [65, 90]   & 77 [67, 84]  & $-$0.31 & 0.20 \\
           & + task       & 99 [97, 100]  & 98 [97, 100] & 0.04 & $-$0.29 \\
\bottomrule
\end{tabular}
\end{table}

\section{Geometry of the representations}
\label{app:geometry}

These analyses test the assumption of Proposition~\ref{prop:coverage}, that the correspondence between two models is close to linear where a map has data. Shape agreement uses the 17 instruction-tuned models of the mapping study (136 pairs), the normalization control our 11 models (110 pairs), and the patch analysis 20 pairs drawn from them; all use the general prompts and the \texttt{qa} template.

\paragraph{Shape agreement.} Table~\ref{tab:app_shape} compares the geometry of the two models' representations of the same prompts without fitting any map: representational similarity (Spearman correlation of the two pairwise distance matrices; RSA), the overlap of each prompt's 10 nearest neighbors (mutual $k$NN), and linear and RBF CKA \citep{kornblith2019similarity}. All four are far above their values for shuffled pairings at every depth, agreement is highest early and lowest at the final block, and models of the same family agree more than models of different families (RSA 0.71 against 0.53; linear CKA 0.87 against 0.82). Pairs whose shapes agree more are fit better by a linear map, with Spearman $\rho$ between RSA and in-domain $R^2$ of 0.58 to 0.67 across depths. The representations are also low-dimensional: their TwoNN intrinsic dimension is 18 to 24 across depths, in spaces with thousands of dimensions.

\begin{table}[h]
\centering
\caption{Shape agreement between models by relative depth (136 pairs), with shuffled-pairing nulls, and the correlation of RSA with linear-map $R^2$.}
\label{tab:app_shape}
\small
\setlength{\tabcolsep}{4pt}
\begin{tabular}{@{}lccccccc@{}}
\toprule
$d$ & RSA & Mutual $k$NN & CKA (linear) & CKA (RBF) & Null RSA & Null CKA & $\rho$(RSA, $R^2$) \\
\midrule
0.25 & 0.77 & 0.44 & 0.87 & 0.91 & 0.00 & 0.01 & 0.58 \\
0.5  & 0.65 & 0.37 & 0.84 & 0.87 & 0.00 & 0.01 & 0.62 \\
0.75 & 0.49 & 0.35 & 0.84 & 0.86 & 0.00 & 0.02 & 0.65 \\
0.9  & 0.46 & 0.35 & 0.85 & 0.86 & 0.00 & 0.03 & 0.66 \\
1    & 0.48 & 0.31 & 0.76 & 0.80 & 0.00 & 0.02 & 0.67 \\
\bottomrule
\end{tabular}
\end{table}

\paragraph{Normalization.} All maps act on representations divided by their norm, which places them on a sphere, a curved space. If the kernel map's advantage came from that curvature, it would shrink when maps are fit on raw activations. Table~\ref{tab:app_norm} shows that it does not: the kernel map gains 0.047 to 0.060 in $R^2$ over the linear map on normalized representations and 0.046 to 0.058 on raw ones, and a linear map fit on raw activations and projected to the sphere does slightly worse than one fit on the sphere directly. The small nonlinearity the kernel map captures is a property of the models' correspondence.

\begin{table}[h]
\centering
\caption{In-domain $R^2$ of global maps on normalized and raw representations (110 pairs). Gaps are kernel minus linear, with 95\% intervals over pairs.}
\label{tab:app_norm}
\small
\setlength{\tabcolsep}{4pt}
\begin{tabular}{@{}lcccccc@{}}
\toprule
 & \multicolumn{2}{c}{Normalized} & \multicolumn{2}{c}{Raw} & \multicolumn{2}{c}{Kernel minus linear} \\
\cmidrule(lr){2-3}\cmidrule(lr){4-5}\cmidrule(l){6-7}
$d$ & Linear & Kernel & Linear & Kernel & Normalized & Raw \\
\midrule
0.25 & 0.600 & 0.647 & 0.601 & 0.647 & 0.047 [0.045, 0.049] & 0.046 [0.044, 0.048] \\
0.5  & 0.447 & 0.500 & 0.449 & 0.501 & 0.053 [0.050, 0.055] & 0.052 [0.049, 0.054] \\
0.75 & 0.360 & 0.417 & 0.359 & 0.415 & 0.056 [0.054, 0.059] & 0.056 [0.054, 0.059] \\
1    & 0.313 & 0.373 & 0.310 & 0.368 & 0.060 [0.058, 0.062] & 0.058 [0.056, 0.060] \\
\bottomrule
\end{tabular}
\end{table}

\paragraph{Local linear patches.} If the correspondence bent strongly, linear maps fit on small neighborhoods would predict better than one global map. For each held-out prompt we fit a linear map on its $k$ nearest training prompts in raw activations, and on $k$ random training prompts as a control (Table~\ref{tab:app_patch}). Neighborhoods help relative to random subsets of the same size, so the correspondence has some local structure. Local maps on 25 to 400 neighbors still predict worse than the global map, and only at 800 to 1,600 neighbors do they match it, improving on it by at most 0.02 in normalized error. Within the region the data cover, a single linear map is about as accurate as a collection of local ones, which is the condition under which Proposition~\ref{prop:coverage} applies.

\begin{table}[h]
\centering
\caption{Normalized prediction error of local linear maps fit on each prompt's $k$ nearest neighbors (raw activations, 20 pairs), against the same map on $k$ random prompts and the global map. Lower is better.}
\label{tab:app_patch}
\small
\setlength{\tabcolsep}{4pt}
\begin{tabular}{@{}lccccccc@{}}
\toprule
 & \multicolumn{2}{c}{$k = 25$} & \multicolumn{2}{c}{$k = 400$} & \multicolumn{2}{c}{$k = 1600$} & \\
\cmidrule(lr){2-3}\cmidrule(lr){4-5}\cmidrule(lr){6-7}
$d$ & Nearest & Random & Nearest & Random & Nearest & Random & Global \\
\midrule
0.25 & 0.273 & 0.510 & 0.167 & 0.258 & 0.155 & 0.195 & 0.163 \\
0.5  & 0.414 & 0.658 & 0.285 & 0.403 & 0.262 & 0.312 & 0.273 \\
0.75 & 0.484 & 0.802 & 0.441 & 0.621 & 0.380 & 0.449 & 0.397 \\
1    & 0.547 & 0.833 & 0.524 & 0.674 & 0.438 & 0.505 & 0.455 \\
\bottomrule
\end{tabular}
\end{table}

\section{Theory and the coverage score}
\label{app:theory}

\paragraph{Setting.} Let $X \in \mathbb{R}^{n \times D_s}$ and $Y \in \mathbb{R}^{n \times D_t}$ hold the source and target representations of the $n$ training prompts as rows, with means $\mu_X, \mu_Y$ and centered versions $X_c, Y_c$. The ridge map is $M(x) = (x - \mu_X) W_\lambda + \mu_Y$ with $W_\lambda = (X_c^\top X_c + \lambda I)^{-1} X_c^\top Y_c$, and $S_\lambda = (X_c^\top X_c + \lambda I)^{-1} X_c^\top X_c$. A tool built from class means has source form $\delta_S = \bar x_A - \bar x_B$ and target form $\delta_T = \bar y_A - \bar y_B$, where the means run over its defining prompts. The correctness direction, an average of per-question differences of class means, has the same form, since it is also a fixed linear combination of representations of the defining prompts. Mapping the defining prompts and taking the difference of their mapped means gives $M$'s constant terms twice with opposite signs, so the carried tool before renormalization is exactly $\delta_S W_\lambda$.

\paragraph{Proof of Proposition~\ref{prop:coverage}.} Under the assumption, every training and defining prompt satisfies $y = xA + b + \varepsilon$ with $\mathbb{E}[\varepsilon \mid x] = 0$. Centering removes $b$, so $Y_c = X_c A + E_c$ with $\mathbb{E}[E_c \mid X] = 0$, and
\begin{equation*}
W_\lambda = S_\lambda A + (X_c^\top X_c + \lambda I)^{-1} X_c^\top E_c, \qquad \mathbb{E}[\delta_S W_\lambda] = \delta_S S_\lambda A .
\end{equation*}
For the native tool, $\delta_T = \delta_S A + (\bar\varepsilon_A - \bar\varepsilon_B)$, so $\mathbb{E}[\delta_T] = \delta_S A$. Since $I - S_\lambda = \lambda (X_c^\top X_c + \lambda I)^{-1}$,
\begin{equation*}
\mathbb{E}[\delta_S W_\lambda] - \mathbb{E}[\delta_T] = -\delta_S (I - S_\lambda) A = -\lambda\, \delta_S (X_c^\top X_c + \lambda I)^{-1} A .
\end{equation*}
For the angle, write $X_c^\top X_c = \sum_i s_i v_i v_i^\top$ over a complete orthonormal basis, with $s_i \ge 0$, and $a_i = \delta_S v_i$. Then
\begin{equation*}
\|\delta_S S_\lambda\|^2 = \sum_i a_i^2 \frac{s_i^2}{(s_i + \lambda)^2}, \qquad \|\delta_S (I - S_\lambda)\|^2 = \sum_i a_i^2 \frac{\lambda^2}{(s_i + \lambda)^2} .
\end{equation*}
Because $s^2 + \lambda^2 \le (s + \lambda)^2$, the two sums add to at most $\|\delta_S\|^2$, so $\|\delta_S (I - S_\lambda)\| \le \|\delta_S\| \sqrt{1 - c^2}$. Let $v = \delta_S A$ be the expected native tool and $e = \delta_S (I - S_\lambda) A$, so that the expected carried tool is $v - e$ and $\|e\| \le \|A\| \|\delta_S\| \sqrt{1 - c^2}$. The sine of the angle between $v$ and $v - e$ is the distance from $v$ to the line through $v - e$, divided by $\|v\|$, which is at most $\|e\| / \|v\|$. Hence $\sin\theta \le \|A\| \|\delta_S\| \sqrt{1 - c^2} / \|\delta_S A\| = \kappa \sqrt{1 - c^2}$. \hfill$\square$

\paragraph{Consequences.} If the training data never vary along the tool, $X_c \delta_S^\top = 0$, every $a_i$ with $s_i > 0$ vanishes, $\delta_S S_\lambda = 0$, and the expected carried tool is zero; this is the behavior of the GSM8K map. The uncovered term $u = \lambda\, \delta_S (X_c^\top X_c + \lambda I)^{-1} \delta_S^\top = \sum_i a_i^2 \lambda / (s_i + \lambda)$ also bounds the error, $\|e\| \le \|A\| \sqrt{u}$, since $\lambda^2/(s+\lambda)^2 \le \lambda/(s+\lambda)$. Adding a prompt $x$ to $n$ training prompts changes the centered scatter matrix exactly to $X_c^\top X_c + \tfrac{n}{n+1}(x - \mu_X)^\top (x - \mu_X)$, a positive semidefinite update, so $(X_c^\top X_c + \lambda I)^{-1}$ cannot increase and $u$ cannot increase at fixed $\lambda$. In our experiments $\lambda$ is re-chosen by leave-one-out error for every map, so the guarantee is approximate there.

\paragraph{Scope.} The proposition compares expected tools in the normalized space where maps are fit, and before the renormalization of mapped representations. The native refusal direction in our experiments is computed from raw activations; the target's own direction computed from normalized representations recovers 96\% of its effect across strengths (Appendix~\ref{app:refusal}), so the comparison is close. The factor $\kappa$ measures how much $A$ shrinks the tool relative to its largest gain and can be large when $A$ is poorly conditioned along the tool; the bound is informative only when $\kappa \sqrt{1 - c^2} < 1$. The proposition makes no claim when the correspondence is not linear on the covered region, which Appendix~\ref{app:geometry} examines empirically.

\paragraph{Computing the score.} The coverage score uses the eigendecomposition already computed to fit the map, in the dual form $X_c X_c^\top = U \operatorname{diag}(s) U^\top$ when $D_s \ge n$: with $g = U^\top X_c \delta_S^\top$, $c^2 = \sum_i g_i^2 s_i / (s_i + \lambda)^2 / \|\delta_S\|^2$, at the $\lambda$ chosen by leave-one-out error. It needs the source representations of the training prompts and of the tool's defining prompts, and the target representations of the training prompts only through the choice of $\lambda$; no target labels enter. We also record two companion measures: the map's $R^2$ on the tool's defining prompts, which needs the target's representations of those prompts but not their labels, and the mean leverage $(x - \mu_X)(X_c^\top X_c + \lambda I)^{-1}(x - \mu_X)^\top$ of the defining prompts.

\paragraph{Results.} Table~\ref{tab:app_cov} gives the correlation of each measure with recovery over all pairs, maps, and seeds, and Table~\ref{tab:app_covarm} the mean coverage and recovery of each map. Coverage and fit on the tool's prompts predict recovery about equally well, and leverage, which grows when the tool's prompts lie far from the training data, predicts it with the opposite sign. The cosine between the carried and native tools predicts best but requires target labels, so it serves only as a check. For a fixed map, coverage does not rank pairs for refusal, TruthfulQA, or sentiment ($|\rho| \le 0.27$ across pairs, under the general map, the smallest coverage level, and the full task set), but it does for ARC with task prompts ($\rho = 0.71$), where larger sources both cover the task better and transfer more (Appendix~\ref{app:pairs}).

\begin{table}[h]
\centering
\caption{Spearman correlation of label-free measures with recovery (sentiment: accuracy gap), over every pair, map, and seed. The cosine to the native tool uses target labels and is shown for reference.}
\label{tab:app_cov}
\small
\setlength{\tabcolsep}{4pt}
\begin{tabular}{@{}lcccccc@{}}
\toprule
Tool & Runs & Coverage & Coverage, within pair & Fit on tool prompts & Leverage & Cosine to native \\
\midrule
Refusal    & 3,500 & 0.67 & 0.75 & 0.74 & $-$0.52 & 0.72 \\
ARC        & 3,690 & 0.71 & 0.73 & 0.67 & $-$0.40 & 0.82 \\
TruthfulQA & 4,510 & 0.68 & 0.80 & 0.67 & $-$0.50 & 0.78 \\
Sentiment  & 4,510 & 0.45 & 0.47 & 0.41 & $-$0.36 & 0.50 \\
\bottomrule
\end{tabular}
\end{table}

\begin{table}[h]
\centering
\caption{Mean coverage score $c$ and mean recovery (\%; sentiment: accuracy gap, points) by map, over pairs and seeds.}
\label{tab:app_covarm}
\small
\setlength{\tabcolsep}{4pt}
\begin{tabular}{@{}lcccccccc@{}}
\toprule
 & \multicolumn{2}{c}{Refusal} & \multicolumn{2}{c}{ARC} & \multicolumn{2}{c}{TruthfulQA} & \multicolumn{2}{c}{Sentiment} \\
\cmidrule(lr){2-3}\cmidrule(lr){4-5}\cmidrule(lr){6-7}\cmidrule(l){8-9}
Map & $c$ & Rec. & $c$ & Rec. & $c$ & Rec. & $c$ & Gap \\
\midrule
Narrow (GSM8K)       & 0.46 & 2   & 0.31 & $-$65 & 0.32 & 15 & 0.30 & $-$32.0 \\
General              & 0.88 & 54  & 0.58 & 16  & 0.64 & 77 & 0.63 & $-$7.8 \\
+ smallest coverage  & 0.97 & 103 & 0.65 & 43  & 0.77 & 90 & 0.67 & $-$4.1 \\
+ other datasets     & 0.96 & 88  & 0.77 & 73  & 0.73 & 82 & 0.87 & $-$0.9 \\
+ task               & 0.995 & 102 & 0.83 & 88 & 0.93 & 98 & 0.90 & 0.2 \\
\bottomrule
\end{tabular}
\end{table}

\section{Ablation with a direction at every block}
\label{app:blocks}

Single-direction ablation removes the direction estimated at the steering depth $d$ from the embeddings and from the output of every block. The per-block variant estimates a separate direction for each of the $L + 1$ positions (the embeddings and the output of each block) and removes each from its own position. For the target's own tool, each direction is the difference of mean raw activations of the 128 harmful and 128 harmless defining prompts at that position. For a carried tool, we fit a task-only map at every position, from the source's position $\operatorname{round}(j L_s / L_t)$ to the target's position $j$, on the 256 harmful and 256 harmless task prompts, and carry the direction through it as in the main experiments. All four variants were run in one job per target with the first seed, so they share prompts and generation settings.

Table~\ref{tab:app_blocks} gives refusal on the held-out harmful prompts. Per-block directions remove more refusal than single directions for both the native and the carried tool, and the carried per-block directions match the native ones: they leave 9\% of harmful prompts refused on average, against 8\%, and recover 101\% $[89, 112]$ of the native per-block effect over the nine targets with coherent native ablations. Single carried directions recover 76\% of the single native direction in the same run. Two targets stand out. In Qwen2.5-1.5B, per-block carried directions leave 40\% of prompts refused, against 6\% for the native ones, so the per-block map does not reach the native effect there. In Phi-3-mini, the native per-block directions do worse than the single native direction, and the carried per-block directions do better than the target's per-block ones but worse than its single direction (Appendix~\ref{app:pairs}). Degenerate outputs are rare for carried directions (0.1\% per-block, 0\% single) and occur mainly in Gemma-2-27B under its own direction, where every response is repetitive in both variants.

\begin{table}[h]
\centering
\caption{Refusal on held-out harmful prompts (\%) under ablation of a single direction or one direction per block, for the target's own tool and for carried tools (mean over 10 sources; task-only maps). $^\ast$All responses degenerate.}
\label{tab:app_blocks}
\small
\setlength{\tabcolsep}{5pt}
\begin{tabular}{@{}lccccc@{}}
\toprule
 & & \multicolumn{2}{c}{Target's own} & \multicolumn{2}{c}{Carried} \\
\cmidrule(lr){3-4}\cmidrule(l){5-6}
Target & No ablation & Single & Per block & Single & Per block \\
\midrule
Llama-3.2-1B  & 73 & 36 & 14 & 47 & 2 \\
Qwen2.5-1.5B  & 91 & 5 & 6 & 39 & 40 \\
Gemma-2-2B    & 94 & 0 & 0 & 8 & 0 \\
Phi-3-mini    & 94 & 12 & 48 & 88 & 33 \\
Qwen2.5-7B    & 89 & 8 & 0 & 37 & 2 \\
Llama-3.1-8B  & 86 & 50 & 0 & 49 & 1 \\
Gemma-2-9B    & 95 & 0 & 0 & 37 & 0 \\
Gemma-2-27B   & 89 & 0$^\ast$ & 0$^\ast$ & 80 & 1 \\
Llama-3.1-70B & 89 & 34 & 0 & 30 & 0 \\
Qwen2.5-72B   & 64 & 23 & 0 & 14 & 2 \\
\midrule
Mean, excluding Gemma-2-27B & 86 & 19 & 8 & 39 & 9 \\
\bottomrule
\end{tabular}
\end{table}

\section{Base models and building tools in the target}
\label{app:base}

\paragraph{Base models.} The four base models of the mapping study are Qwen2.5-7B, Llama-3.1-8B, Gemma-2-9B, and Mistral-7B (v0.3) before instruction tuning. Their general activations come from the mapping study, and we extract correctness and sentiment states with the same templates and depth as for the instruction-tuned models. Refusal is not tested, because base models do not refuse reliably. The same inclusion criteria apply: on ARC only the Qwen2.5-7B and Gemma-2-9B base models beat zero-shot scoring by at least 5 points (50\% to 72\% and 63\% to 77\%), while all four pass on TruthfulQA (28\% to 29\% zero-shot, 70\% to 75\% native) and sentiment (68\% to 91\%). Table~\ref{tab:app_base} gives mean recovery by kind of pair, using the first seed. With task prompts in the map, TruthfulQA and sentiment transfer in every direction, including between a base model and its own instruction-tuned version, while the general map transfers the sentiment probe least between base models and transfers ARC poorly into base models. Sentiment recovery here is measured against chance, as in Equation~\ref{eq:recovery} with $E_0 = 50\%$.

\begin{table}[h]
\centering
\caption{Transfer between base and instruction-tuned models: mean recovery (\%) by kind of pair (first seed). ``Own'' pairs a base model with its instruction-tuned version; $n$ is the number of pairs.}
\label{tab:app_base}
\small
\setlength{\tabcolsep}{4pt}
\begin{tabular}{@{}llcccc@{}}
\toprule
Tool & Pairs & $n$ & General & General + task & Mapped random \\
\midrule
ARC        & base $\to$ base           & 6  & $-$9 & 61 & $-$190 \\
           & base $\to$ instruct       & 32 & 24 & 88 & $-$135 \\
           & base $\to$ own instruct   & 4  & 44 & 80 & $-$132 \\
           & instruct $\to$ base       & 20 & $-$63 & 68 & $-$185 \\
           & instruct $\to$ own base   & 2  & 34 & 92 & $-$171 \\
\addlinespace
TruthfulQA & base $\to$ base           & 12 & 89 & 100 & 8 \\
           & base $\to$ instruct       & 40 & 72 & 97 & $-$3 \\
           & base $\to$ own instruct   & 4  & 85 & 99 & $-$19 \\
           & instruct $\to$ base       & 40 & 78 & 101 & 17 \\
           & instruct $\to$ own base   & 4  & 98 & 101 & 23 \\
\addlinespace
Sentiment  & base $\to$ base           & 12 & 48 & 97 & $-$9 \\
           & base $\to$ instruct       & 40 & 81 & 93 & $-$7 \\
           & base $\to$ own instruct   & 4  & 90 & 101 & $-$3 \\
           & instruct $\to$ base       & 40 & 66 & 93 & $-$11 \\
           & instruct $\to$ own base   & 4  & 107 & 97 & $-$13 \\
\bottomrule
\end{tabular}
\end{table}

\label{app:rebuild}
\paragraph{Building the tool in the target.} Table~\ref{tab:app_rebuild} extends Table~\ref{tab:rebuild} to every number of prompts, over seeds 1 to 4. For correctness, the target's direction built from $k$ labeled questions takes the first $k$ of the seed's 250-question pool, so at $k = 50$ it is the native direction by construction and above 50 it uses more labels than the native one. The unsupervised direction is the top principal component of the answer candidates of the same $k$ questions the map sees, after centering the candidates within each question, with its sign chosen so that its top-scoring candidates agree more often with zero-shot scoring; the oracle sign is the better of the two. For sentiment, the unsupervised probe is the top principal component of the same reviews, thresholded at their median projection, with the better of the two signs. For refusal, the target's own direction is the difference of means of the same task prompts the map sees, and the unsupervised direction is the top principal component of those prompts pooled, with its sign set by which class projects higher.

\begin{table}[h]
\centering
\caption{Carried tools against tools built in the target from the same $k$ prompts (seeds 1 to 4; recovery, \%, except sentiment: accuracy minus the target's own probe, points). Intervals are in Table~\ref{tab:rebuild} for the rows shown there.}
\label{tab:app_rebuild}
\small
\setlength{\tabcolsep}{4pt}
\begin{tabular}{@{}llccccc@{}}
\toprule
 & & & \multicolumn{3}{c}{Built in the target} \\
\cmidrule(l){4-6}
Tool & $k$ & Carried & PC, zero-shot sign & PC, oracle sign & Labeled \\
\midrule
Refusal (per class) & 16  & 103 & \multicolumn{2}{c}{92 (sign from categories)} & 97 \\
                    & 32  & 103 & -- & -- & 95 \\
                    & 64  & 102 & -- & -- & 97 \\
                    & 256 & 102 & \multicolumn{2}{c}{97 (sign from categories)} & 100 \\
\addlinespace
ARC (questions)     & 10  & 44 & $-$76 & $-$67 & 70 \\
                    & 25  & 65 & $-$43 & $-$29 & 86 \\
                    & 50  & 78 & $-$18 & $-$16 & 100 \\
                    & 100 & 84 & $-$7  & $-$7  & 105 \\
                    & 200 & 88 & $-$14 & $-$9  & 109 \\
\addlinespace
TruthfulQA (questions) & 10  & 91 & 81 & 84 & 96 \\
                       & 25  & 93 & 72 & 86 & 99 \\
                       & 50  & 95 & 86 & 86 & 100 \\
                       & 100 & 97 & 86 & 86 & 100 \\
                       & 200 & 98 & 86 & 86 & 101 \\
\addlinespace
Sentiment (reviews) & 10  & $-$3.6 & -- & $-$22.3 & $-$12.7 \\
                    & 25  & $-$3.7 & -- & $-$22.3 & $-$2.5 \\
                    & 50  & $-$0.5 & -- & $-$20.3 & $-$0.5 \\
                    & 100 & $+$0.2 & -- & $-$22.3 & $-$1.0 \\
                    & 256 & $+$0.8 & -- & $-$22.3 & $+$1.6 \\
\bottomrule
\end{tabular}
\end{table}

\end{document}